\documentclass[a4paper]{article}

\usepackage{INTERSPEECH2020}
\usepackage{amsmath}
\usepackage{multirow}
\usepackage{amsfonts}
\usepackage{setspace}
\title{Style Variation as a Vantage Point for Code-Switching}
\name{Khyathi Raghavi Chandu, Alan W Black}
\address{
  Language Technologies Institute \\ Carnegie Mellon University}
\email{kchandu@cs.cmu.edu, awb@cs.cmu.edu}

\begin{document}

\maketitle
\begin{abstract}
  Code-Switching (CS) is a common phenomenon observed in several bilingual and multilingual communities, thereby attaining prevalence in digital and social media platforms. This increasing prominence demands the need to model CS languages for critical downstream tasks. A major problem in this domain is the dearth of annotated data and a substantial corpora to train large scale neural models. Generating vast amounts of quality text assists several down stream tasks that heavily rely on language modeling such as speech recognition, text-to-speech synthesis etc,. We present a novel vantage point of CS to be style variations between both the participating languages. Our approach does not need any external annotations such as lexical language ids. It mainly relies on easily obtainable monolingual corpora without any parallel alignment and a limited set of naturally CS sentences. We propose a two-stage generative adversarial training approach where the first stage generates competitive negative examples for CS and the second stage generates more realistic CS sentences. 
We present our experiments on the following pairs of languages: Spanish-English, Mandarin-English, Hindi-English and Arabic-French. We show that the trends in metrics for generated CS move closer to real CS data in each of the above language pairs through the dual stage training process. We believe this viewpoint of CS as style variations opens new perspectives for modeling various tasks in CS text.
\end{abstract}

\section{Introduction}
Code-Switched \cite{poplack1978syntactic} text is prevalent in semi-formal and informal communication platforms. 
A major challenge in addressing this widely observed form of mixing languages is the scarcity of curated data, thereby making it a low resource setting \cite{sitaram2019survey}. However, there are plenty of monolingual corpora available for each of the participating languages. We present a novel standpoint to transfer knowledge from monolingual corpora without additional annotations such as language ids or parse trees. The recent advances in cross-lingual pretrained language models \cite{artetxe2019massively, conneau2019cross} call out for vast amounts of code-switched data. Hence, our work on automatic generation of CS text is relevant for several downstream tasks.

We propose a novel vantage point for CS to be observed as stylistic variation between the participating embedded and matrix languages. For the scope of this paper, we define the style variations between languages to be extrinsic properties such as surface lexical forms and intrinsic properties such as underlying grammar, word order etc,. We address this problem with adversarial training in two stages: (1) \textit{Stage 1:}  transfer the style of each of the monolingual participating languages into the content of the other language; (2) \textit{Stage 2: } discriminating between the incorrectly switched and naturally switched sentences. The four styles in play here are the following: (1) $l_{m}$: matrix language style (2) $l_{e}$: embedded language style (3) $l_{a}$: incorrect/artificial code-switching style (4) $l_{n}$: natural code-switching style. The goal is to traverse smoothly across these styles without affecting the content. The first stage generates negative examples facilitating the discriminative training for the second stage. This dual stage training eliminates the need for additional linguistic annotations, such as language id used by several contemporary works. We present our results on four pairs of languages.

\section{Related Work}

\begin{figure*}[t!]
\centering
\includegraphics[trim={1cm 3.5cm 2cm 2.2cm},clip,width=0.94\linewidth]{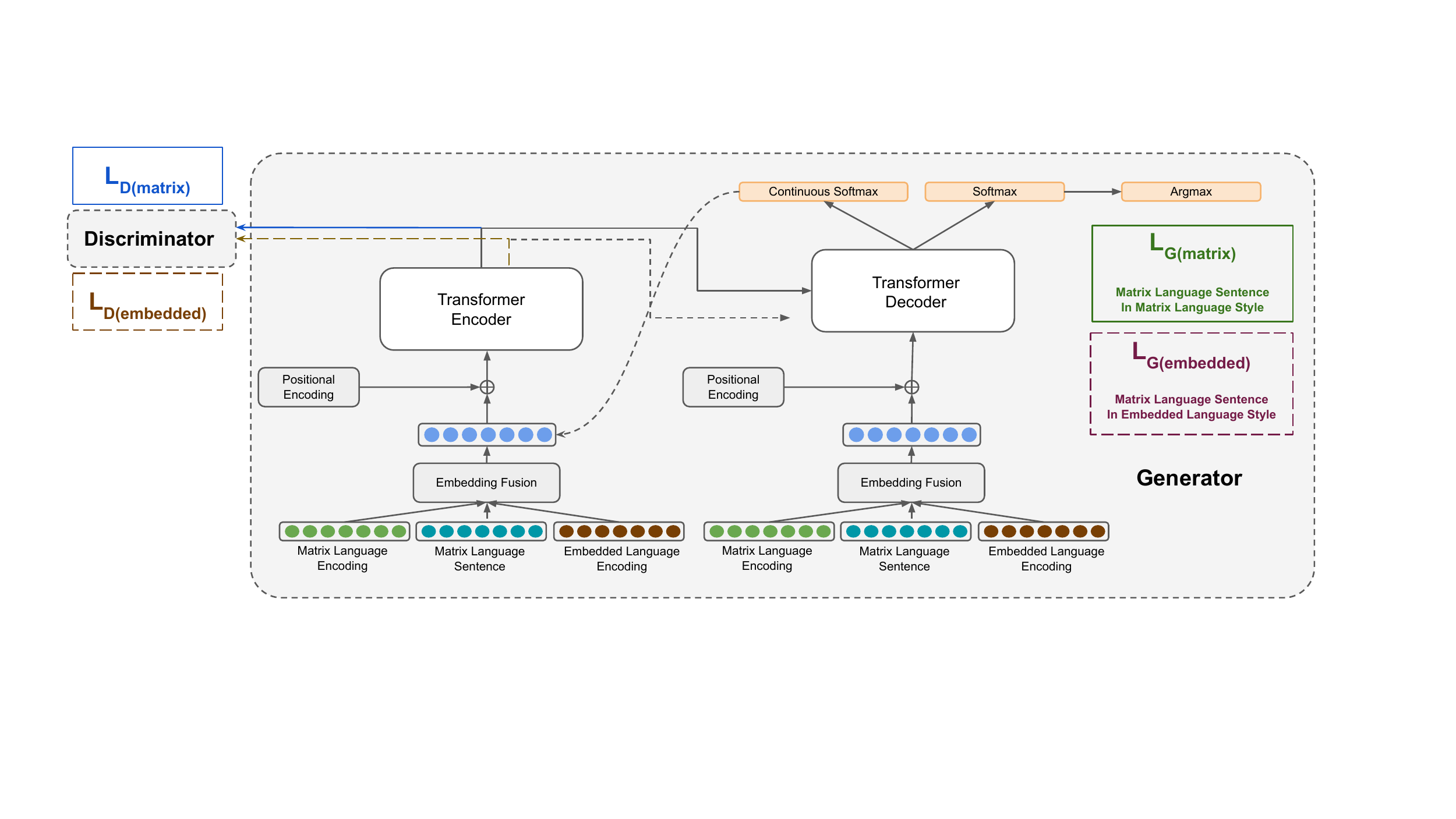}
\caption{ {\small Transformer based GAN architecture for generating CS text. \textit{Note: The same architecture is used for two stages. Matrix language sentence is the embedding of the text and language encoding is the embedding of the language.}} }
\label{fig:overall_preposal}
\end{figure*}

\noindent \textbf{Constraint Theory based Generation:} \cite{li2012code} combined syntactic constraints by predicting language boundary to reconstruct CS text. \cite{pratapa2018language} and \cite{lee2019linguistically} present techniques based on Equivalence Theory \cite{poplack1980sometimes} and Matrix Language Frame
Theory \cite{myers1997duelling} to create grammatically valid CS text. While these methods demonstrate the use of expert knowledge to assist generation, the same is difficult to replicate and scale to other languages. 

\noindent \textbf{Language Informed Modeling: } Prior works rely on annotations of language spans in multi-task setup \cite{chandu2018language} or using dual RNN to handle each language \cite{garg2018code}. They generated sentences which are used to pretrain the model which essentially is augmenting the original data with generated data. In contrast, our approach uses the generated data to discriminate against the real CS data in Stage 2 to enhance the generation. As a result, \cite{garg2018code} trains on CS data to generate new CS text. Our approach relies primarily on a lot of monolingual data in Stage 1 and some amount of CS data. Capturing syntactic and language switching signals prove effective in a hierarchical VAE architecture \cite{samanta2019deep}. \cite{chang2018code} proposed a GAN based approach to generate language id tags and discriminate whether it is a valid sequence. The similarity between this and our works is that the fundamental architecture is a GAN. Lexical level translation is needed by \cite{chang2018code} which cannot be done by a simple word lookup but depends on the context. The language id based lookup for translation may not perform well in all cases especially when transliteration is also needed.

\noindent \textbf{Dependence on Parallel data: } \cite{winata2019code} proposed a seq-to-seq model with copy mechanism limiting the method to rely on parallel monolingual translations of CS text. However, more often there might be limitations on gathering parallel data not only across languages but in this case also with CS text. There has been some prior work in style transfer techniques for non-parallel data such as \cite{yang2018unsupervised, shen2017style}.

Our approach eliminates the need for drafting constraint theories, additional annotations for language ids and parallel data. This enables scalability to new language pairs attributed to the availability of monolingual corpora and limited CS text. 

\section{Datasets}

Each of the participating monolingual utterances (belonging to $\mathbb{M}$ (matrix language) and $\mathbb{E}$ (embedded language)) are treated as two distinct styles. Note that the sentences are not aligned either at phrase or sentence levels. We explored CS for four language pairs as presented in Table \ref{tab:datasets}.

The reasons behind selecting these language pairs are multi-step. We selected Hinglish and Spanglish since they are widely spoken languages. The usage of Hindi in Hinglish is commonly romanized, bringing in a new variety to the platform. Spanglish and Hinglish thus have very close scripts as opposed to Mandarin-English where the scripts are different. While the word order for English, Spanish, Mandarin and French is subject-verb-object (SVO), the same for Hindi is subject-object-verb (SOV) and Arabic is verb-subject-object (VSO). These differences facilitate the stylistic attributes to the mixing of these languages.

\begin{table}[h]
\scriptsize
\centering
\begin{tabular}{lll}
\hline
\textbf{Language} & \textbf{Monolingual} & \textbf{Code-Switched}\\
\hline
Spanish & \cite{graff2010fisher} & \multirow{2}{*}{\cite{deuchar2014building} }\\
English & \cite{budzianowski2018multiwoz} & \\
\hline
Mandarin & \cite{tian2014corpus} & \multirow{2}{*} {\cite{lyu2010seame}} \\ 
English & \cite{budzianowski2018multiwoz} & \\
\hline
Hindi & \cite{mathur2018detecting} & \multirow{2}{*}{\cite{mathur2018detecting}}\\
English & \cite{mathur2018detecting} &  \\
\hline
Arabic & \cite{song2014collecting} & \multirow{2}{*}{\cite{cotterell2014algerian}}\\
French & \cite{koehn2005europarl} & \\
\hline
\end{tabular}
\caption{Monolingual and Code-Switched Datasets used for training Stage 1 and Stage 2}
\label{tab:datasets}
\end{table}


\section{Model Description}


The two problems we address are repealing the need for annotations (such as language id) on CS data and maximizing the utilization of monolingual data. Both the issues are addressed using a two stage generative adversarial training paradigm with a transformer based autoencoder. Unavailability of parallel sentences is tackled by preserving semantics of the original sentence of one language and mixing the attributes of the other language without disentangling the representation into these two properties. Following are the two stages involved: \\
\noindent \textbf{Stage 1}: The embedded and matrix languages are mixed in arbitrary ways to generate CS text. This stage simply uses the corpora from each language as an individual style.\\
\noindent \textbf{Stage 2}: The sentences generated after Stage 1 were not supervised via any real CS sentences. Hence, they are used as negative examples (with style $l_a$) against limited amount of CS text  (with style $l_n$) to generate naturally switched sentences.

The architecture remains same for both stages except for variation in hyperparameters.  Figure \ref{fig:overall_preposal} presents our GAN setup for Stage 1. The following subsections present the flow by instantiating for Stage 1 for readability. The same process is applied for Stage 2 with the difference of using positive and negative examples of CS sentences.


\subsection{Generator}

The generator in our architecture comprises of transformer based encoder and decoder. In Stage 1, our transformer encoder takes in the matrix language sentence ($s_{m}$ $\in$ $\mathbb{M}$ ) along with the matrix language encoding or style ($l_{m}$) and produces a latent representation ($z_{m,m}$).
\setlength{\belowdisplayskip}{0pt}
\setlength{\abovedisplayskip}{0pt}
\begin{equation}
\small
    z_{m,m} = TransEnc(s_{m}, l_{m}) \forall s_m \in \mathbb{M} 
\end{equation}

\noindent We use this $z_{m,m}$ along with the original matrix language sentence $s_{m}$ and the matrix language encoding $l_{m}$ to reconstruct the original sentence $s_{m}$. Greedy decoding is performed that uses argmax which is non-differentiable to compute the loss for reconstructing the original sentence ($L_{G(matrix)}$). 
\begin{equation}
\small
    L_{G(matrix)} = - \sum\limits_{s_m \in \mathbb{M}}log (Pr(s_m | s_m, l=m)) 
\end{equation}
\noindent Next, the same matrix language sentence $s_{m}$ is considered along with the embedded language encoding $l_{e}$ to produce a latent representation ($z_{m,e}$).
\begin{equation}
\small
    z_{m,e} = TransEnc(s_{m}, l_{e}) \forall s_m \in \mathbb{M} 
\end{equation}

\noindent This $z_{m,e}$ is used to reconstruct the original sentence $s_{m}$. This means that the model is attempting to reconstruct the content of the original sentence while varying the style i.e, language encoding. The loss corresponding to this reconstruction is ($L_{G(embedded)}$).
\begin{equation}
\small
    L_{G(embedded)} = - \sum\limits_{s_m \in \mathbb{M}}log (Pr(s_m | s_m, l=e)) 
\end{equation}

Similarly, corresponding counterparts using $z_{e,e}$ and $z_{e,m}$ are generated.

\begin{figure*}[t!]
\centering
\includegraphics[trim={0cm 9.2cm 0cm 5.9cm},clip,width=0.94\linewidth]{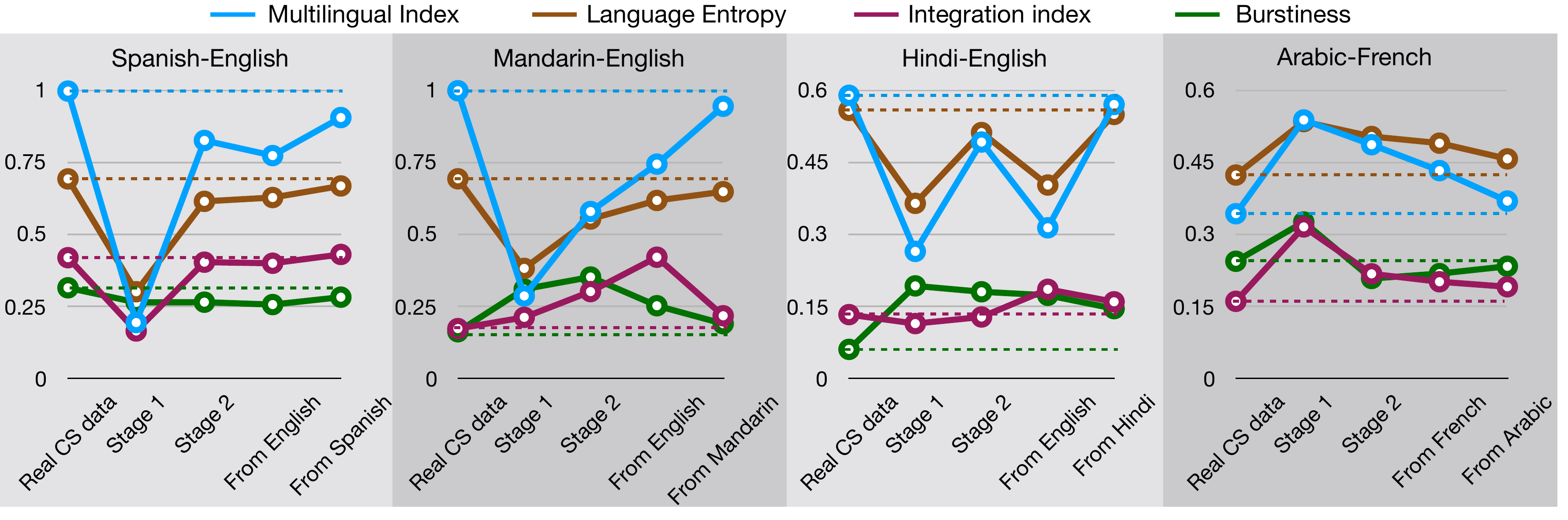}
\caption{ Trends in metrics for evaluating the generation of CS text for four language pairs in dual stage training. \small {\textit{The dotted line of each color benchmarks the corresponding metric for real CS data.} }}
\label{fig:results}
\end{figure*}

\subsection{Discriminator}

The discriminator is a classifier that predicts whether the current distribution is closer to the original latent space or the generated latent space. 
The purpose of generator reconstructing the original sentence $s_{m}$ with matrix language encoding $l_{m}$ (contributing to $L_{G(matrix)}$) is solely to make sure that the generator is retaining the content of the original sentence, and has no contribution towards training the discriminator. On the other hand, the generation of the sentence $s_{m}$ with embedded language encoding $l_{e}$, say $s_{m,e}$ essentially establishes our end goal. In the GAN architecture, we now have two choices i.e, sampling a sentence from: (1) original distribution $s_{m,m}$ i.e, the matrix language sentence with the matrix language encoding or (2) distribution from the generator $s_{m,e}$ i.e, the matrix language sentence with the embedded language encoding.  The positive examples to train the discriminator come from real sentences which are trained by maximizing the probability for predicting that it belongs to label \textit{m}. 
\begin{equation}
\small
\begin{split}
    L_{D(matrix)} = - \sum log (Pr(m | z_{m,m}, l=m )) 
\end{split}
\end{equation}

One particular problem for training GANs in text domain is the non-differentiable function of argmax that is performed in decoding. There are three prominent solutions to address this problem including REINFORCE \cite{williams1992simple}, Gumbel-Softmax \cite{jang2016categorical}, manipulating the latent space. We proceed with the third option by performing a continuous softmax of the words, thus eliminating the need to perform argmax, which is described in detail here. Let the vocab size be $\mathcal{V}$ and the embedding dimension be $\mathcal{H}$. Instead of discretely making a selection of the embedding over the vocabulary space to select each word, we perform continuous softmax. The final softmax layer in decoder provides us with a vector of size $1 \times \mathcal{V}$. Multiplying this with the embedding weights ($\mathcal{V} \times \mathcal{H}$) results in $1 \times \mathcal{H}$ vectors for each word. Note that in the case of argmax, we make a discrete selection of the word, whereas, in the case of continuous softmax, we arrive at a soft representation of the weighted combination of properties of the words across different words in the vocabulary. Therefore the latter does not enforce this soft representation to be a word. This partially decoded representation now passes through the transformer encoder to arrive at a latent representation to be fed into the discriminator. 
\begin{equation}
\small
    L_{D(embedded)} = - \sum log (Pr(e | z_{m,e}, l=e ))
\end{equation}

\subsection{Dual Stage Training Setup: }

The task of generating CS text not only entails mixing languages but also mixing them appropriately. This means that our discriminator performs two tasks of discriminating between: (1) the participating languages, owing to the asymmetry between their interactions, such as matrix and embedded languages (Stage 1) and (2) incorrectly and correctly switched languages (Stage 2). Hence we dissolve the training procedure into two stages with each stage dedicated to one of the aforementioned tasks.

In Stage 1, the sentences from $s_m \in \mathbb{M}$ are transferred to the style of $l_e$ ($s_{m,e}$). At any given point of time, there is only one sentence and one language style that is encoded as presented in equations 1 and 3. We conflated this information while presenting in Figure \ref{fig:overall_preposal}. For example, let the subscript `r' be reconstructed sentence from an original source sentence. In addition to the sentence, there is one style that is taken as input. This style could be either (only one among) the matrix language or the embedded language style which is encoded. Let the source and the matrix language styles be $s_m$ and $s_e$ respectively. The discriminator labels the following representations in the corresponding ways:
(i) Label 0: (a) $s_m$ with $l_{m}$ style; (b) $`s_{mr}'$ with $l_{m}$ style;
(ii) Label 1: (a) $s_e$ with $l_{e}$ style; (b) $`s_{er}'$ with $l_{e}$ style;
In this way, we train a model to generate $s_{m,e}$ and $s_{e,m}$ which are matrix language sentences in the embedded language style and embedded language sentences in the matrix language style.
Since there is no supervision from naturally CS sentences, we delegate this responsibility to the second stage of training with the same architecture. We used $s_{m,e}$ as negative examples of CS sentences for Stage 2 of training. We have also experimented with a random subset of $s_{m,e}$ and $s_{e,m}$ as negative examples. This performed worse than the former setting since $s_{m,e}$ has the underlying grammatical structure of $\mathbb{M}$, thereby generating stronger negative examples for adversarial training. In Stage 2, the negative examples from generated above belongs to incorrect CS style because the words are mixed arbitrarily from both the languages. The style from real CS data is the correct style in which we want to finally generate CS text. The same model is trained in the same way described above again with these two styles in this second stage. Note that the training objective is also the same in the second stage as the first stage. The goal of the first stage is to generate robust negative examples of CS style to train the model in the second stage.

\noindent \textbf{Hyperparameter setup:} We used 3 layers of transformer encoders and decoders with a maximum sequence length of 45 words. The word embedding dimension is 256 with 300 iterations of pre-training the generator before training our GAN.

In Stage 1, there is minimal overlap of vocabulary between the languages. This is in contrary to data in typical style transfer datasets which have overlapping vocabulary spaces. Hence the discriminator learns much faster than generator in our case. To combat this, the learning rate in Stage 1 for the generator is 1e-3 and the discriminator is 1e-4. For Stage 2, the generator and discriminator are initialized with models learnt in Stage 1 thereby transferring the knowledge of each of the languages. However, to quickly adapt to the parameter space of Stage 2, we use slanted triangular learning rate \cite{howard2018universal} with a short linear increase period followed by longer decay period. Adam optimizers are used throughout the model. We plan to release our code, models and generated samples upon acceptance.



\section{Evaluation}

We present our results on four pairs of languages from Table \ref{tab:datasets}. We evaluate trends in different  metrics of CS proposed by \cite{guzman2017metrics} in our dual stage training. Consolidated results are presented in Figure \ref{fig:results}. Metrics that we look into are \textit{multilingual-index}, \textit{language entropy},  \textit{integration-index} and \textit{burstiness}. In Figure \ref{fig:results}, `Stage 1' contains generated sentences $s_{m,e}$. The model learnt in Stage 2 has options to generate from negative examples or original text of each language as source. `Stage 2' contains sentences generated using negative examples from Stage 1 as source. Similarly, `From \textit{$<$lang$>$}' uses the corresponding language as source. We observe that the metrics move closer to real CS in `Stage 2' as compared to `Stage 1'. Within `Stage 2', metrics are closer to real CS data when the source text belongs to $\mathbb{M}$ in comparison to $\mathbb{E}$ or $s_{m,e}$ from Stage 1. We plan to explore properties of syntactic and semantic mixing in each stage in our future work. 

\section{Qualitative Analysis}

The following are some of the common forms of errors observed in the generated text for Hinglish when trained on the blogging data collected from \cite{chandu2018language}.
\begin{itemize}
    \item \textit{Gender Disagreement:}  
    For instance, consider the sentence `kyunki ye scam bhi ho sakti hai' (Meaning: because this can also be a scam). The gender of the direct object which is `scam' should agree with the inflection of the verb `sakti'. Hence this should have been `sakta'. The gender of the word `scam' (which is a borrowed word from embedded language English) is unknown in the matrix language, so it would be presumed to be masculine. But this sentence used a feminine verb phrase.
    
    \item \textit{Incorrect Case markers:}
    `agar aap bhi ye post pasand aaye toh aap puch sakte hai' (Meaning: If this post is pleasing to you also, then you can ask). The words `to you' when used in Hindi is supposed to be in dative case which is `aap ko' in the first clause of the sentence. However `aap' which is in the nominative case is generated.
    
    \item \textit{Semantically incorrect due to random mixing:} 
    Sometimes, the model also generate completely random mix of words that semantically are incorrect. For instance, this is one of the sentence from the output: `am bahut hi achchi jankari aapko pata hi hoga' (Loosely Translated Meaning: very good information you must be knowing). The sentence does not convey a coherent meaning. The word order of the sentence is also jumbled and does not strictly belong to either of the matrix or the embedded languages.
    
    \item \textit{No mixing: } In some cases, the entire sentence is built from words belonging to the same language. For instance, 
    `the best way to improve your application for the reasons listed below'.
    
    \item \textit{Incorrect sub-word mixing: } Though the incorrect case markers and gender disagreement are syntactic errors, this seems to happen due to the modeling at word level. Sub-word level modeling of text is a promising direction to address this error category especially for morphologically rich languages.
    
\end{itemize}

The category of first two error types described above are syntactic. Our current model is purely data driven from the surface forms. This motivates the utility of inducing syntax while generation. The semantically incorrect sentences seem to be generated due to random mixing of the words from both the languages. The loosely translated meaning of the sentence is not utterly senseless but when framed in CS fashion does not make sense. In addition, there is an inherent challenge while dealing with multiple datasets that lead upto domain variation. For instance, the vocabulary or the style on social media platforms such as Twitter is very different from the domain of conversations. Although we have carefully selected the datasets to belong to similar domains, this might often not be feasible which invites the domain invariant modeling of text.



\section{Conclusion}

We present a novel perspective of viewing CS as style variants between participating languages. We believe this viewpoint opens new avenues for dealing with mixed language text. The main contributions of the paper are threefold. Firstly, we eliminate the need for explicit language identification using two stage adversarial training. Secondly, our approach transfers from bountiful monolingual resources and relies on limited CS data to generate new CS sentences. Thirdly, we present our experiments on dual stage transformer based GAN model for generating four pairs of CS languages: Spanish-English, Mandarin-English, Hindi-English and Arabic-French. 
In future, we would like to compare the performance of this technique with other style transfer models and also explore the possibility of end-to-end training of both the stages.


\bibliographystyle{IEEEtran}

\bibliography{mybib}


\end{document}